**Title**

Automated Classification of Tutors' Dialogue Acts Using Generative AI: A Case Study Using the CIMA Corpus

**Authors**

*1/ Liqun He (Corresponding)*

Institution: University College London (Gower Street, London, WC1E 6BT)

Email: liqun.he@ucl.ac.uk

ORCiD: 0000-0002-0837-5857

*2/ Jiaqi Xu*

Institution: Zhejiang University (866 Yuhangtang Rd, Hangzhou, 310027)

Email: jiaqi.xu@zju.edu.cn

**Abstract**

This study explores the use of generative AI for automating the classification of tutors' Dialogue Acts (DAs), aiming to reduce the time and effort required by traditional manual coding. This case study uses the open-source CIMA corpus, in which tutors' responses are pre-annotated into four DA categories. Both GPT-3.5-turbo and GPT-4 models were tested using tailored prompts. Results show that GPT-4 achieved 80% accuracy, a weighted F1-score of 0.81, and a Cohen's Kappa of 0.74, surpassing baseline performance and indicating substantial agreement with human annotations. These findings suggest that generative AI has strong potential to provide an efficient and accessible approach to DA classification, with meaningful implications for educational dialogue analysis. The study also highlights the importance of task-specific label definitions and contextual information in enhancing the quality of automated annotation. Finally, it underscores the ethical considerations associated with the use of generative AI and the need for responsible and transparent research practices. The script of this research is publicly available at https://github.com/liqunhe27/Generative-AI-for-educational-dialogue-act-tagging.

# 1 Introduction

In educational settings, dialogue plays a crucial role. It is through dialogue that educators can convey and achieve various pedagogical goals (Chi et al., 2001; Roscoe & Chi, 2007). Therefore, a systematic description and analysis of the tutor-student dialogue is necessary to gain a detailed understanding of the tutoring process and improve it further. Traditional methods of analysing dialogue heavily rely on manual coding, in which human coders observe the transcriptions and assign label(s) for each speaking turn or utterance based on predetermined categories (i.e., the coding scheme) (Huxley, 2020). However, this approach is immensely complex and cognitively demanding, which hinders the understanding and analysis of teaching processes (Hennessy et al., 2020).

To automate the human coding process and maximise dialogue analysis for improving teaching, some previous research used natural language processing (NLP) techniques to train automatic coding models. Although some achievements have been made in this area

(Lin et al., 2022; Song et al., 2021; Stasaski et al., 2020), these attempts still require humans to label a part of the data to train the model for automatically labelling other parts of the data set. Due to the need for manual pre-annotation and technical expertise in model training, current automatic coding methods are not widely used by educators and researchers.

To address these current limitations, this study aims to investigate the potential of generative AI for classifying tutors' DAs, with an emphasis on pursuing satisfactory performance by developing more effective prompts. Given that generative AI enables a more natural and direct approach to utilising models, this study aims to contribute to an easier-to-use automatic coding method for educators and researchers to facilitate their analysis of tutoring processes, ultimately improving teaching and learning. Additionally, the scripts have been made openly available to facilitate further research.

## 2   Literature Review

### 2.1   Tutoring Dialogue Analysis

Dialogue is the main tool for teaching in educational settings. Tutor intentions are conveyed and reflected by what they say and when they say it (Chi et al., 2001) and through various forms of dialogue, tutors achieve various pedagogical goals (Roscoe & Chi, 2007). Thus, a systematic examination of the tutoring dialogue between teachers and students is necessary for understanding and enhancing the quality of online tutoring (Chen et al., 2021; Lin et al., 2022). Since we are uttering, we are not just making sounds, but aiming to achieve our intention behind the utterances (Searle, 1965, 1969) .Thus, in analysing tutoring dialogue, the emphasis lies not only on what is said (i.e. the form of utterances), but on why it is said

(the function of the utterances). 'Dialogue act' (DA), also known as 'speech act', is used to represent the hidden intention of the functioning of the utterance (Traum, 1999).

According to Speech Act theory (Searle, 1965, 1969), identifying tutors' DAs requires inferring and categorising based on what the tutor is saying. The traditional approach to achieve it is through human experts' observation and manual coding. Through manually classifying DAs, we can capture key characteristics of tutors' interactions during tutoring processes and enable subsequent systematic and quantitative analysis of tutoring processes (Maharjan et al., 2018). However, since each tutorial session may contain numerous DAs, relying solely on human coding is immensely complex and cognitively demanding (Hennessy et al., 2020).

## 2.2 Two Paradigms for Automatic DA Coding

Automating the coding process is thus a natural solution to maximise the benefits of systematic coding and reduce the significant time required for human coding. As noted by Stolcke et al. (2000), automating the detection and modelling of DAs is crucial for comprehending dialogue.

In the field of NLP, the machine (automatic) coding process is related to the 'text classification task' and there are currently two pre-trained large language models (PLLM)-based paradigms for achieving this - pre-training-*fine-tuning* paradigm and pre-training-*prompting* paradigm (Zhao et al., 2023). Both paradigms initially rely on pre-training large language models on massive unlabelled corpuses (e.g., Wikipedia), giving them the preliminary understanding of human natural language structures and information (Che & Liu, 2022).Through pre-training, models can transform human natural language into a computer-understandable format. Subsequently, to perform DA classification, the *fine-tuning* paradigm further trains (i.e., fine-tunes) models on additional smaller labelled corpuses to help them further learn how to perform specific tasks and deliver expected outputs (Che & Liu, 2022). However, fine-tuning requires labelling corpus data and a substantial technical foundation, making this method difficult for educators and researchers to implement.

Recently, advances in NLP have led to a new *prompting* paradigm. Using a method known as Reinforcement Learning from Human Feedback (RLHF) (Ouyang et al., 2022), PLLMs have gained the ability to interact with humans in natural language and respond without requiring further fine-tuning (Zhao et al., 2023). These PLLMs, such as ChatGPT, are referred to as instruction-based PLLMs or generative AI, with the message that elicits a model response being called a 'prompt'. For example, you can input a natural language prompt to the model, like 'What is the capital of France?'. Without additional training, the model can directly respond with something like 'Paris'. This advancement has significantly enhanced the user-friendliness of PLLMs. Anyone, regardless of their prior machine learning expertise, can achieve results within a few minutes, which was previously impossible. To maximise the value of instruction-based PLLMs, we need to further examine how to better design and develop the prompts that guide them.

## 2.3 Related Works on Automatic DA Coding

Since instruction-based PLLMs have recently emerged, research in this area is relatively limited. Previous studies have primarily focused on fine-tuning PLLMs for automating DA coding. Specifically, previous attempts usually employed PLLMs like Global Vectors for Word Representation (GloVe) (Pennington et al., 2014) and Bidirectional Encoder Representations from Transformers (BERT) (Devlin et al., 2019). All of these models are designed to help machines translate human language into a format they can understand. Subsequently, through fine-tuning, machines can ultimately acquire the ability to automatically identify and label tutors' verbal actions. For instance, Stasaski et al. (2020) achieved an F1 score (which is a metric for evaluating the quality of model performance in a task, with values closer to 1 indicating better performance) of 0.72 using a GloVe-based model on a dataset of 2,296 samples. Song et al. (2021) trained a BERT-based model to classify classroom dialogue into seven categories, achieving an F1 score of 0.68 on a dataset of 15K labelled turns. Additionally, Lin et al. (2022) developed BERT-based model to attain an F1 score of 0.74 for automatically identifying and tagging 31 various tutoring actions.

However, the fine-tuning paradigm requires manual data annotation and model fine-tuning, significantly increasing the complexity and cost of automatic coding. Meanwhile, due to the novelty of the prompting paradigm, there is limited research on the use of generative AI in coding tutoring dialogues. These constraints have motivated us to explore the potential of generative AI in automatic coding, with the aim of providing a more user-friendly and cost-effective solution for analysing tutoring processes, ultimately leading to improvements in real-world tutoring practices.

## 3  Methods

### 3.1  Dataset

The dataset used in this study is the 'Prepositional Phrases' sub-dataset of the Conversational Instruction with Multi-responses and Actions (CIMA) open-source dataset (Stasaski et al., 2020), which contains 1,135 labelled tutor replies and their prior context data around teaching Italian vocabulary. This dataset was collected through role-playing by crowd workers pretending to be tutors and students. The tutor role-players would read the previous tutor-student dialogue. Following this, they would provide response to the students' responses and label the response at the same time. Most responses will be categorised into four categories, namely 'Question,' 'Hint,' 'Correction,' and 'Confirmation,' with any few unclassifiable ones being categorised as 'Others.' Specifically:

- **Question**: The tutor asks the student a question to check understanding or extend the conversation
- **Hint**: The tutor scaffolds a student's understanding through providing hints
- **Correction**: The tutor corrects a mistake or misconception made by the student
- **Confirmation**: The tutor confirms a student's answer or understanding is correct

The advantage of using pre-annotated open datasets is that it allows for quick algorithm development and iteration, without having to initially invest substantial time in data collection and annotation (Maayan, 2019), which well fits the purpose of our study. This dataset is made available under the Creative Commons 2.5 license.

3.2  Prompts

A 'prompt' refers to the input given to the generative AI model to solicit its response. Usually, each prompt consists of two components: the 'system message', which specifies what role the AI should play and how it should behave generally, and the 'user message', which is the specific user's question or comment the AI responds to.

To achieve satisfactory performance, we initially investigated how various system message affect the performance of generative AI in automatically categorising (i.e., coding) tutors' actions. Specifically, we developed four different system messages to prompt the model:

*Table 1 - Four System Messages for Prompting Generative AI in Classifying Tutors' DAs*

| Number | System Message (Bolded part for emphasis on differences between instructions) |
|---|---|
| **1 (basic)** | You will be given a snippet of a tutor-student conversation enclosed within triple backticks.<br><br>Your job is to carefully read this information line by line, and then provide a tag of the TUTOR_RESPONSE from the following list.<br><br>Choose ONLY ONE best tag from the list of tags provided here.<br><br>- Question<br>- Hint<br>- Correction<br>- Confirmation |
| **2 (elaborative)** | You will be given a snippet of a tutor-student conversation enclosed within triple backticks.<br><br>Your job is to carefully read this information line by line, and then provide a tag of the TUTOR_RESPONSE from the following list.<br><br>**After the colon, each tag's explanation is attached.**<br><br>Choose ONLY ONE best tag from the list of tags provided here.<br><br>- Question: The tutor is asking an open-ended question usually with a question mark<br>- Hint: The tutor is answering a student's question and/or providing additional information<br>- Correction: The tutor is explicitly correcting student's mistake<br>- Confirmation: The tutor is agreeing with student's words |
| **3 (chain-of-thought)** | You will be given a snippet of a tutor-student conversation enclosed within triple backticks, |

|  | **Your job is to perform the following steps:** |
|---|---|
|  | 1. Identify student's utterance and tutor's response. |
|  | 2. Determine whether the tutor action is 'Question', 'Confirmation', or none of them ('Other'). |
|  | 3. Unless it's NOT 'Other', choose ONLY ONE tag from 'Question' and 'Confirmation' and skip other steps. If it's 'Other', proceed to step 4. |
|  | 4. Determine whether the tutor action is 'Hint' or 'Correction'. |
|  | Choose ONLY ONE tag from 'Hint' and 'Correction' |
| **4 (combined)** | You will be given a snippet of a tutor-student conversation enclosed within triple backticks. |
|  | Your job is to perform the following steps: |
|  | 1. Identify student's utterance and tutor's response. |
|  | 2. Determine whether the tutor is asking an open-ended question usually with a question mark ('Question'), agreeing with student's words ('Confirmation'), or none of them ('Other'). |
|  | 3. Unless it's NOT 'Other', choose ONLY ONE tag from 'Question' and 'Confirmation' and skip other steps. If it's 'Other', proceed to step 4. |
|  | 4. Determine whether tutor is answering a student's question and/or providing additional information ('Hint') or is explicitly correcting student's mistake ('Correction'). |
|  | Choose ONLY ONE tag from 'Hint' and 'Correction' |

In Table 1, we illustrated four different system messages for prompting generative AI to classify tutors' DAs. Specifically, Prompt 1 (Basic) is the foundational version, where we simply asked the model to read the information we provided and predict which category this response belongs to - Questions, Hint, Correction, or Confirmation. This prompt is developed mainly followed the example of Review classifier provided by OpenAI platform (OpenAI,

2023). Prompt 2 (elaborative) built upon Prompt 1 by providing additional information about the definitions of these labels while keeping other aspects the same. Prompt 3 (chain-of-thought), another extension of Prompt 1, provided detailed step-by-step instructions for guiding the model, while other elements remain unchanged. This encourages generative AI to respond in steps. Prompt 4 (combined) brought together features from both Prompts 2 and 3, offering explanations for each tag and providing step-by-step instructions.

Prior research has also shown that providing contextual conversation could improve the model performance of identifying tutors' DA (Caines et al., 2022). Thus, in this study, we also tested different user messages with a focus on incorporating prior contextual conversations. Specifically, we examined three scenarios ('n' is used to represent the number of previous conversational turns):

1) **n=0:** inputting only the tutor's response without contextual information,
2) **n=1:** inputting the tutor's response with the previous student's utterance, and
3) **n=2:** inputting the tutor's response with the previous round of tutor-student exchange.

Correspondingly, we made changes in system messages for these three situations. Specifically, in the 2nd and 4th system messages, we modified the expression of the first step, 'Identify student's utterance and tutor's response', as follows:

- for n=0, it's now 'Identify tutor's response,' and
- for n=2, it's 'Identify tutor's initiation, student's utterance, and tutor's response.'

### 3.3 Experimental Procedures

As for the data pre-processing, we initially excluded tutors' response categorised as 'others' due to their ambiguous nature and the very small proportion (less than 1% of all responses).

This exclusion reduced the number of samples from 1,135 to 1,065. As the dataset is crowd-sourced, multiple role-player tutors may respond to the same student's turn. To ensure sample diversity, we randomly selected only one response from multiple responses. A total of 80 samples were randomly selected from this dataset for further analysis, 20 samples from each label. After this pre-processing, each sample contains the previous round of tutor-student exchange, the tutor's response, and the human annotations.

The model was then implemented using Application Programming Interface (API). The entire study comprised two main phrases. In the first phase, we used the 'gpt-3.5-turbo' model with four prompt types (basic, elaborate, chain-of-thought, combined). Each prompt was tested with 0, 1, and 2 preceding turn(s) as input, resulting in a total of 12 experiments from all combinations. In each experiment, model performance was evaluated using F1 scores for each label and the weighted average F1 score.

In the second phrase, we tested the top five performing combinations from the 12 experiments using the 'gpt-4' model to evaluate its further potential. We did not test all combinations with gpt-4 for two main reasons: (1) at the time of the experiments in 2023, gpt-4 had a stricter speed limit at 10,000 tokens per minute, reducing experimental efficiency, and (2) our study primarily aims to explore the potential of generative AI in machine coding tutoring DAs, rather than investigating the impact of different models. In addition to recording F1 scores for each label and weighted average F1 scores, this phrase also included the measurement of accuracy (A), precision (P), recall (R), and Cohen's K.

# 4 Results

The results of the first phase are presented in *Table 2*, with the top five combinations highlighted in bold.

*Table 2 - Performance of the gpt-3.5-turbo Model for Machine Coding Tutors' DAs Under Different Conditions (Top 5 scores are bolded)*

| Experimental Conditions | f1-score | | | | |
| --- | --- | --- | --- | --- | --- |
| | Confirmation | Correction | Hint | Question | avg. |
| 1 basic / n=0 | 0.41 | 0.51 | 0.00 | 0.62 | 0.39 |
| 1 basic / n=1 | 0.85 | 0.58 | 0.16 | 0.46 | 0.51 |
| 1 basic / n=2 | 0.74 | 0.58 | 0.15 | 0.40 | 0.47 |
| **2 elaborative / n=0** | **0.67** | **0.62** | **0.16** | **0.70** | **0.54** |
| **2 elaborative / n=1** | **0.81** | **0.80** | **0.65** | **0.79** | **0.76** |
| **2 elaborative / n=2** | **0.79** | **0.84** | **0.72** | **0.67** | **0.76** |
| 3 chain-of-thought / n=0 | 0.63 | 0.63 | 0.00 | 0.71 | 0.49 |
| 3 chain-of-thought / n=1 | 0.62 | 0.78 | 0.00 | 0.55 | 0.49 |
| 3 chain-of-thought / n=2 | 0.70 | 0.62 | 0.00 | 0.53 | 0.46 |
| **4 combined / n=0** | **0.74** | **0.63** | **0.46** | **0.64** | **0.62** |
| **4 combined / n=1** | **0.82** | **0.89** | **0.61** | **0.72** | **0.76** |
| **4 combined / n=2** | **0.61** | **0.55** | **0.53** | **0.70** | **0.60** |

To analyse the impact of different system messages on model performance, we averaged the results of three scenarios using 0, 1, and 2 preceding turns for each system message (as displayed in *Table 3*). This allows us to represent the average model performance under each system message, as evaluated by the F1 score. The results show that providing additional information about label definitions significantly enhances performance. In both cases where system messages included additional label definition information (prompt 2 and 4, bolded in the table), the weighted average F1 scores are higher than those without this information (prompt 1 and 3). Besides, the chain-of-thought strategy alone (prompt 3) did not yield a particularly significant improvement.

*Table 3 - Impact of System Messages on Model Performance*

| Experimental Conditions | f1-score | | | | |
|---|---|---|---|---|---|
| | Confirmation | Correction | Hint | Question | avg. |
| 1 Basic | 0.67 | 0.56 | 0.10 | 0.49 | 0.46 |
| **2 elaborative** | **0.76** | **0.75** | **0.51** | **0.72** | **0.69** |
| 3 chain-of-thought | 0.65 | 0.68 | 0.00 | 0.60 | 0.48 |
| **4 combined** | **0.72** | **0.69** | **0.53** | **0.69** | **0.66** |

The *Table 4* displays the impact of providing different previous conversations on model performance. For each scenario, we averaged the results of four different system messages. The results indicate that incorporating contextual conversation in user messages leads to higher weighted F1 scores compared to solely relying on the target turn for identification. This finding is consistent with previous experiments based on fine-tuning DistilBERT (Caines et al., 2022), demonstrating the importance of context in tagging tutors' DAs. However, we found that providing one preceding turn (n = 1) resulted in better average performance than providing more context (n = 2). A possible reason is that, for gpt-3.5-turbo, additional information may introduce unnecessary noise.

*Table 4 - Impact of Contextual Conversation on Model Performance*

| Experimental Conditions | f1-score | | | | |
|---|---|---|---|---|---|
| | Confirmation | Correction | Hint | Question | avg. |
| n=0 | 0.61 | 0.60 | 0.16 | 0.67 | 0.51 |
| **n=1** | **0.78** | **0.76** | **0.36** | **0.63** | **0.63** |
| n=2 | 0.71 | 0.65 | 0.35 | 0.58 | 0.57 |

The results of the second phase are presented in *Table 5*. In this phase, we conducted experiments to evaluate the performance of the 'gpt-4' model under the top five performing conditional combinations identified in the first phase. The results show that under the best-performing condition (utilising a combined system message with the previous tutor-student exchange), we were able to accurately assign 80% labels automatically to tutors' DAs for the tested samples with a weighted F1-score of 0.81. Furthermore, Cohen's Kappa yielded a value of 0.74, indicating substantial agreement between the machine coder's predictions and the human annotations.

*Table 5 - Performance of 'gpt-4' Model Under Top Five Conditional Combinations*

| Experimental Conditions | f1-score | | | | | Cohen's k | A | P | R |
|---|---|---|---|---|---|---|---|---|---|
| | Confirm. | Correction | Hint | Question | avg. | | | | |
| 4 combined / n=1 | 0.81 | 0.79 | 0.70 | 0.70 | 0.75 | 0.67 | 0.75 | 0.76 | 0.75 |
| 2 elaborative / n=2 | 0.84 | 0.84 | 0.72 | 0.63 | 0.76 | 0.68 | 0.76 | 0.76 | 0.76 |
| 2 elaborative / n=1 | 0.81 | 0.78 | 0.56 | 0.61 | 0.69 | 0.60 | 0.70 | 0.71 | 0.70 |
| 4 combined / n=0 | 0.86 | 0.67 | 0.43 | 0.80 | 0.69 | 0.59 | 0.69 | 0.69 | 0.69 |
| **4 combined / n=2** | **0.88** | **0.86** | **0.77** | **0.71** | **0.81** | **0.74** | **0.80** | **0.80** | **0.80** |

# 5 Discussion and Conclusion

## 5.1 Effectiveness of Generative AI for Dialogue Act Coding

This study demonstrates that generative AI has strong potential for automatically classifying tutors' DAs in an accurate and accessible manner. Using a combined prompt that includes clear label definitions, step-by-step instructions, and two preceding tutor-student turns (n = 2), the GPT-4 model achieved 80% accuracy in categorising tutors' DAs into four categories. A Cohen's Kappa value of 0.74 further indicates substantial agreement between machine coding and human coding. Compared to a GloVe-based baseline model, this approach achieved improved performance, increasing F1 scores from 0.72 to 0.81.

More importantly, these improvements were achieved by using natural language instructions to elicit the models' responses, without the need for human annotation or additional training (fine-tuning). This substantially reduced model training time and costs, and simplified model development and implementation, advancing the potential of machine coding in tutoring dialogue analysis.

This study also emphasised the importance of well-defined categories and contextual information. When compared to the original prompt (prompt 1), we found that by including specific definitions of each label in system messages, it can enhance the accuracy of the model's predictions, aligning it more closely with human coding. This improvement may be due to the PLLM model being primarily pre-trained on general web text, providing only a basic understanding of labels; however, for coding educational DAs, task-specific clarification is essential. Besides, the model performance has been further enhanced by providing contextual conversation (i.e., 1 or 2 preceding turns before the target turn) in user messages. This underscores the pivotal role of contextualisation in identifying DAs. This indicates that accurately understanding the intention behind what is said depends not only on the target utterance but also on its surrounding context. As an example from everyday life, the utterance 'I am hungry' can indicate a desire for lunch if spoken by a friend but could convey a different intention, such as a request for money, if spoken by a beggar.

Additionally, it's worth noting that, despite previous studies suggesting that guiding generative AI through a step-by-step process, i.e., using Chain of Thought (CoT) prompting, can lead to better performance in arithmetic and symbolic reasoning tasks (Kojima et al., 2023; Wei et al., 2023), this advantage was not evident in this study for the DA tagging task. As shown in *Table 3*, when all other conditions remained the same, using stepwise instructions (prompt 3 compared to prompt 1, and prompt 4 compared to prompt 2) did not lead to a significant improvement and, at times, even resulted in a decline. However, when using the gpt-4 model (*Table 5*), there was some improvement with CoT. Due to the limited size of the experimental dataset, further investigation is needed to explore this aspect.

## 5.2 Implications

The generative AI-based method proposed here offers a more efficient and user-friendly approach to analysing educational dialogues. It eliminates the need for pre-annotation and the technical complexities associated with traditional machine coding methods, enabling educators and researchers to rapidly conduct educational dialogue research, gaining deeper insights into the dynamics of the teaching and learning process, and identifying effective interactional behaviours for improvement.

Furthermore, the satisfactory coding agreement between AI and human coders also suggests the potential for generative AI to serve as a human *co-coder*, enhancing the quality of educational DA coding. Traditionally, human coding usually requires at least two trained coders to independently code dialogue transcriptions, with a third educational expert resolving disagreements if necessary (Hennessy et al., 2016; Lin et al., 2022). The satisfactory performance of a generative AI coder makes it a potential co-coder, streamlining the human data tagging process and supporting more rigorous educational dialogue analysis.

## 5.3 Methodological Limitations

Firstly, this study only employed a relatively small dataset (n = 80). This decision was primarily driven by the research's main goal, which was to investigate the feasibility of using generative AI for automatically coding tutoring dialogue actions, while also accounting for the current limitations in model speed. In future research endeavours, employing a larger dataset is imperative to further test and validate the conclusions drawn in this study.

Furthermore, only a four-category DA coding scheme was tested. To gain a more detailed understanding of tutoring dialogue processes, dialogue analysis often requires more complex coding schemes with a greater number of categories. Thus, future research should investigate generative AI's performance in educational DA classification tasks that involve more categories.

The response time is another aspect that needs to be considered in the future. Although generative AI does not require additional training time, the time it takes to generate responses can be relatively long. In real-world settings, tutorials often involve several hundred dialogue turns or even more. Therefore, it's crucial to record and consider the response time of machine coding in future research and recognise it as a significant factor in practical applications.

Finally, this study was conducted in 2023, prior to the release of more advanced models such as GPT-4o and Deepseek-R1. As a result, comparisons with these newer models were not included. Future research should evaluate the performance of similar tasks using these more recent models.

5.4  Ethical Considerations

Beyond the methodological limitations outlined earlier, the use of generative AI tools such as ChatGPT also introduces ethical concerns.

Informed Consent. The use of generative AI in data analysis introduces complexities in obtaining truly informed consent. As Stahl & Eke (2024) argue, meaningful consent requires participants to fully understand the purpose, implications, and risks of data use, rather than agreeing to vague terms. Thus, when introducing generative AI as a potential annotation tool, researchers must clearly explain its role, *all* possible uses, and associated risks. This helps ensure that all participants (e.g., learners and tutors) are fully aware of the AI's involvement before giving consent (Wright, 2011).

Privacy and Data Protection. Submitting research data, such as users' dialogue transcripts, to external generative AI tools like ChatGPT raises concerns regarding data reuse and potential third-party sharing, posing risks to participant privacy. For example, submitted data may be used to train or improve AI models (Morgan, 2023) or be accessed by external parties (Stahl & Eke, 2024). To mitigate these risks, researchers should follow data minimisation principles by collecting only essential personal information. Additionally, anonymisation or pseudonymisation should be employed to remove or mask any identifiers that may reveal individual identities (Wright, 2011).

Transparency and Bias. Another key challenge is the limited transparency of AI systems. Particularly, many generative AI providers do not clearly disclose the details of their training data (Morgan, 2023). This lack of transparency can reduce public trust and confidence (Wright, 2011) and makes it difficult to identify and address potential biases within the data, such as those related to race, gender, or sexuality (Morgan, 2023). Researchers must acknowledge these limitations and communicate them clearly to participants, while also advocating for stronger regulatory frameworks to ensure the ethical use of AI technologies (Stahl & Eke, 2024).

## 5.5  Conclusion

This study demonstrates the promising potential of generative AI for automated coding of tutors' DAs, achieving 80% accuracy and substantial agreement with human coders (Cohen's Kappa = 0.74). These results outperform baseline methods while eliminating the need for manual annotation or model fine-tuning, thereby reducing time and cost. The findings also highlight the importance of task-specific label definitions and contextual information in enhancing model performance. As a practical and accessible alternative to traditional coding methods, this approach enables more efficient educational dialogues analysis and offers potential as a reliable human co-coder. However, the study is limited by its small dataset and simple four-category coding scheme. Future research should validate these results using larger datasets and more complex classification tasks. Lastly, the use of generative AI introduces important ethical considerations, which must be addressed through responsible research practices.

# Abbreviations

**API** Application Programming Interface

**BERT** Bidirectional Encoder Representations from Transformers

**CIMA** Conversational Instruction with Multi-responses and Actions

**CoT** Chain of Thought

**DA** Dialogue Act

**GPT** Generative Pre-trained Transformer

**NLP** Natural Language Processing

**PLLM** Pre-trained Large Language Model

**RLHF** Reinforcement Learning from Human Feedback

# Funding

The author(s) received no financial support for the research, authorship, and/or publication of this article.

# Declarations and conflict of interests

All efforts to sufficiently blind the authors during peer review of this article have been made. The authors declare no further conflicts with this work.